\documentclass[runningheads]{llncs}
\usepackage{booktabs}

\usepackage[T1]{fontenc}
\usepackage{graphicx,verbatim}
\usepackage{hyperref}
\usepackage{orcidlink}
\usepackage{booktabs}%
\usepackage{multirow}%
\usepackage{amsmath,amssymb,amsfonts}%
\usepackage{cite}
\usepackage{subcaption}
\usepackage{lineno}
\usepackage[normalem]{ulem}
\usepackage[table]{xcolor}

\usepackage[misc,geometry]{ifsym}
\usepackage{orcidlink}
\newcommand{\repeatthanks}{\textsuperscript{\thefootnote}}

\begin{document}
\title{Surgical Anatomy Recognition with Context Learning using Foundation Representations}
\titlerunning{Anatomy Recognition using Foundation Representations}
%

\author{
Ronald L. P. D. de Jong\inst{1}\thanks{R. L. P. D. de Jong and T. J. M. Jaspers contributed equally to this work.}$^\textrm{(\Letter)}$\orcidlink{0009-0005-7806-4340}
\and
Tim J. M. Jaspers\inst{2}\repeatthanks$^\textrm{(\Letter)}$\orcidlink{0009-0001-8306-5058}
\and
Raf A. H. Vervoort\inst{2}\orcidlink{0009-0001-1587-9324}
\and
Aron F. H. A. Bakker\inst{4,5}\orcidlink{0000-0001-7852-9332}
\and
Yiping Li\inst{1}\orcidlink{0009-0005-0239-3682}
\and
Jip L. Tolenaar\inst{3}\orcidlink{0009-0004-1974-6846}
\and
Jelle P. Ruurda\inst{3}\orcidlink{0000-0001-6584-1677}
\and
Willem M. Brinkman\inst{4}\orcidlink{0000-0001-7883-0213}
\and
Josien P. W. Pluim\inst{1}\orcidlink{0000-0001-7327-9178}
\and
Marcel Breeuwer\inst{1}\orcidlink{0000-0003-1822-8970}
\and
Daan de Geus\inst{6}\orcidlink{0000-0003-0559-5341}
\and
Fons van der Sommen\inst{2}\orcidlink{0000-0002-3593-2356}
}

\authorrunning{R. L. P. D. de Jong et al.}

\institute{Department of Biomedical Engineering, Medical Image Analysis, Eindhoven University of Technology, Eindhoven, The Netherlands\\ \email{r.l.p.d.d.jong@tue.nl} \and Department of Electrical Engineering, Architectures for Reliable Image Analysis, Eindhoven University of Technology, Eindhoven, The Netherlands\\ \email{t.j.m.jaspers@tue.nl} \and Department of Surgery, University Medical Center Utrecht, Utrecht, The Netherlands \and Department of Oncological Urology, University Medical Center Utrecht, Utrecht, The Netherlands \and Department of Urology, Catharina Hospital, Eindhoven, The Netherlands \and Department of Electrical Engineering,  Mobile Perception Systems Lab, Eindhoven University of Technology, Eindhoven, The Netherlands}
  
\maketitle

\begin{abstract}
Accurate recognition of anatomical structures is essential for safe and effective minimally invasive surgery (MIS), yet it remains underexplored in surgical computer vision due to limited annotated data and methods tailored primarily to natural scenes. In this work, we present a combined dataset and model framework to advance anatomy-aware perception in MIS. First, we introduce ATLAS-120k, a large-scale clip-level semantic segmentation dataset comprising over 120,000 annotated frames from 100 surgical videos spanning 14 procedures and multiple modalities, including laparoscopic and robot-assisted surgery. The dataset captures substantial procedural variability and was created using a scalable annotation pipeline that integrates expert manual labeling, automated propagation, iterative refinement, and surgeon verification to ensure high-quality annotations. Second, we propose ATLAS (\underline{A}natomy Recognition with Contex\underline{t} \underline{L}earning using Found\underline{a}tion Representation\underline{s}), a video semantic segmentation model specifically designed for surgical anatomy recognition. Unlike conventional approaches that emphasize object tracking, ATLAS leverages foundation-model embeddings together with lightweight temporal reasoning to incorporate contextual cues such as procedure type, surgical phase, and short-term visual memory. This design enables temporally consistent and accurate predictions while maintaining real-time feasibility. Together, the dataset and model establish a practical foundation for robust surgical scene understanding and support the development of clinically applicable guidance systems for minimally invasive surgery. The models, dataset annotations and annotation platform are publicly available at: \url{https://github.com/TimJaspers0801/ATLAS}.   


\end{abstract}
\section{Introduction}
\label{sec:introduction}
In recent years the field of surgical computer vision has advanced rapidly, driven by breakthroughs in deep learning and image analysis. These methods promise tangible benefits for minimally invasive surgery (MIS), from real-time guidance to improved precision and patient safety, yet clinical translation remains limited. A key bottleneck is the lack of large, diverse, and richly annotated datasets that capture the variability of real surgical practice and thus enable robust, generalizable models. This scarcity has constrained progress despite growing interest in anatomy-aware guidance systems~\cite{MAIERHEIN2022, denBoer2023, dejong2025bench}.

Most prior work has concentrated on well-defined tasks such as surgical phase recognition~\cite{heichole, Cholec80}, tool segmentation~\cite{roboticinstrumentsegmentation}, and instrument tracking~\cite{surgloc}. While these tasks are important building blocks for computer-assisted surgery, accurate recognition of anatomical structures is both a distinct and underexplored challenge: in MIS, surgeons must reason about anatomy with limited viewpoints and without tactile feedback, so robust anatomical perception is essential for safe navigation. Existing anatomy datasets have begun to address this need, but many suffer from limited procedural diversity, relatively few videos or frames, and incomplete coverage of anatomical classes, factors that limit model generalization in realistic settings~\cite{dsad, Cholecseg8k, Mascagni2025}.

To address these gaps we contribute two complementary advances. First, we introduce ATLAS-120k, a large clip-level surgical anatomy segmentation dataset that substantially broadens procedural and technological diversity: it comprises annotations drawn from 100 videos spanning 14 distinct procedures, and includes both laparoscopic and robot-assisted MIS, resulting in over 120k annotated frames. 

Second, we present ATLAS (\underline{A}natomy Recognition with Contex\underline{t} \underline{L}earning using Found\underline{a}tion Representation\underline{s}), a video semantic segmentation model designed specifically for surgical anatomy recognition. Surgical anatomy segmentation, unlike video segmentation for most natural scenes, depends heavily on contextual cues that surgeons use in practice. Specifically, the type of procedure and surgical phase (i.e., a short, distinct stage of the surgery) are critical to understand which anatomical categories may appear. ATLAS integrates lightweight temporal and tracking components to capture this procedure- and phase-dependent context and leverages strong embeddings from a surgical foundation model~\cite{jaspers2025surgenet} to further enhance its domain knowledge. The resulting architecture enables practical real-time operation, while substantially improving anatomical consistency and accuracy across clips and procedures.

Together, ATLAS-120k and the ATLAS model provide a paired dataset and method package designed to advance anatomy-aware surgical perception: the dataset expands the scope and realism of available supervision, and the model demonstrates how foundation-level representations plus compact temporal reasoning can deliver accurate, real-time anatomical segmentation across a wide range of MIS procedures. These contributions aim to close the gap between research prototypes and clinically useful, anatomy-aware guidance systems and represent a step toward safer, more intelligent assistance in the operating room.

\section{Methods}
\begin{figure}[!t]
    \centering
    \includegraphics[width=\linewidth]{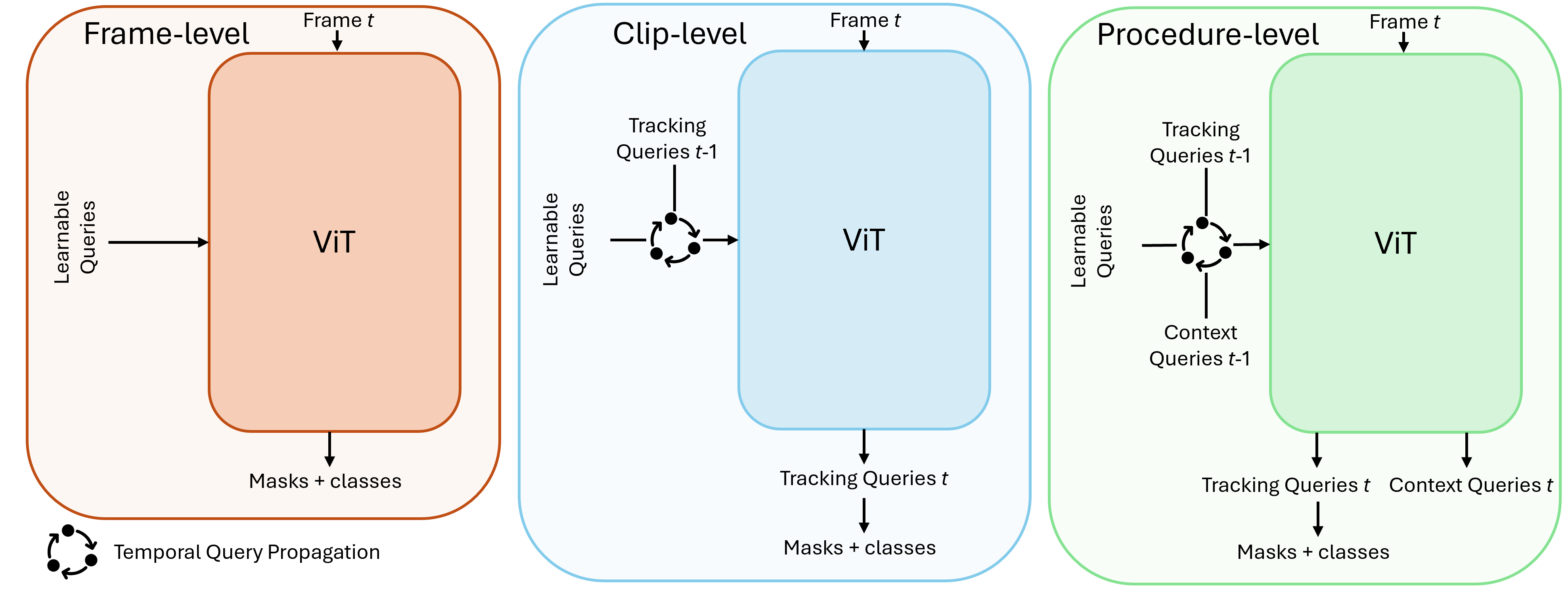}
    \caption{Left: the frame-level semantic segmentation model. Middle: the clip-level model with integrated tracking queries. Right: the procedure-level model, incorporating context queries.}
    \label{fig:method}
\end{figure}

\textbf{Overview.}
Video semantic segmentation in surgical settings presents unique challenges compared to natural videos. While tracking objects across frames is important, successful surgical segmentation also requires domain-specific knowledge: surgeons rely on procedural understanding and awareness of the current surgical phase to identify anatomical structures accurately. Capturing this contextual information is therefore critical.

We build on EoMT~\cite{eomt}, a state-of-the-art image segmentation model that leverages strong pretrained visual embeddings. EoMT uses a ViT encoder to process learnable queries, each representing a single object, and predicts a class and segmentation mask for each of these queries---conducting dense segmentation without complex decoders. VidEoMT~\cite{videomt} extends EoMT to videos by propagating queries from frame $t-1$ to frame $t$, allowing segmentation and tracking over time. However, surgical videos also require knowledge of procedure and phase, which VidEoMT does not capture. To address this, we introduce context queries that augment the segmentation and tracking queries. These queries encode global procedural and temporal information, encouraging the model to integrate prior knowledge with temporal context for more accurate and consistent segmentation. Figure~\ref{fig:method} provides an overview of the proposed method.

\textbf{Transformer Input.}
For each video frame $t$, the input to the transformer consists of
\begin{equation}
    X_t = (P_t,\; Q^\textrm{seg}_{0},\; Q^\textrm{proc}_{0},\; Q^\textrm{phase}_{0},\; \hat{Q}^\textrm{seg}_{t-1},\; \hat{Q}^\textrm{proc}_{t-1},\; \hat{Q}^\textrm{phase}_{t-1} ),
\end{equation}
where $P_t$ are visual patch tokens, $Q^\textrm{seg}_{0}$ are learnable segmentation queries, and $\hat{Q}^\textrm{seg}_{t-1}$ are propagated segmentation queries from the previous frame. We additionally introduce learnable initial context queries $Q^\textrm{proc}_{0}$ and $Q^\textrm{phase}_{0}$ that are refined over time. Previous-frame context queries ($t\!-\!1$) provide temporal continuity. 

All queries are processed jointly by a single transformer encoder. Because attention is global, segmentation queries can directly attend to both visual evidence and contextual queries, enabling context-aware predictions.

\textbf{Procedure Context Queries.}
Procedure queries encode high-level surgical context that remains relatively stable across a video. For each time step, the updated procedure query embeddings $\hat Q^\textrm{proc}_t$ are passed through a multilayer perceptron (MLP) to predict a procedure label
\begin{equation}
\hat y^\textrm{proc}_t = \text{MLP}(\hat Q^\textrm{proc}_t).
\end{equation}
We compute the cross-entropy loss between $\hat y^\textrm{proc}_t$ and ground-truth label $y^\textrm{proc}_t$:
\begin{equation}
\mathcal{L}_\textrm{proc} = \text{CE}(\hat y^\textrm{proc}_t, y^\textrm{proc}_t).
\end{equation}
This auxiliary supervision encourages the procedure queries to represent global scene context useful for segmentation, such as instrument presence or anatomical exposure patterns.

\textbf{Phase Context Queries.}
Phase queries capture finer temporal variations. For each clip, we obtain a phase embedding
\begin{equation}
z_t = \text{MLP}(\hat Q^\textrm{phase}_t).
\end{equation}
We apply an InfoNCE loss that induces an attractive force between embeddings from temporally adjacent frames within the same clip ($z_{t^+}$), while repelling embeddings from different clips ($z_{j}$) or distant timestamps using scaling factor $\tau$:
\begin{equation}
\mathcal{L}_\textrm{phase} = -\log \frac{\exp(z_t \cdot z_{t^+}/\tau)}
{\sum_j \exp(z_t \cdot z_j/\tau)}.
\end{equation}
This self-supervised signal encourages phase queries to encode temporally coherent surgical state information.

\textbf{Temporal Query Propagation.}
To ensure short-term memory and temporal consistency, queries are propagated over time in a manner similar to VidEoMT~\cite{videomt}. The input queries $Q_t$ for the encoder-only model at frame $t$ are a function of the output queries $\hat{Q}_{t-1}$ of the model at frame $t-1$ and the initial learnable queries $Q_0$:
\begin{equation}
Q_t = \mathrm{Linear}(\hat{Q}_{t-1}) + Q_0,
\end{equation}
where $\mathrm{Linear}$ is a linear layer and
\begin{equation}
\hat{Q}_{t-1} = 
\begin{bmatrix}
\hat{Q}^\textrm{seg}_{t-1} & \hat{Q}^\textrm{proc}_{t-1} & \hat{Q}^\textrm{phase}_{t-1}
\end{bmatrix}
\text{, and }
Q_0 = 
\begin{bmatrix}
Q^\textrm{seg}_0 & Q^\textrm{proc}_0 & Q^\textrm{phase}_0
\end{bmatrix}.
\end{equation}

\section{Anatomy Segmentation Dataset}

\begin{figure}[!t]
    \centering
    \includegraphics[width=\linewidth]{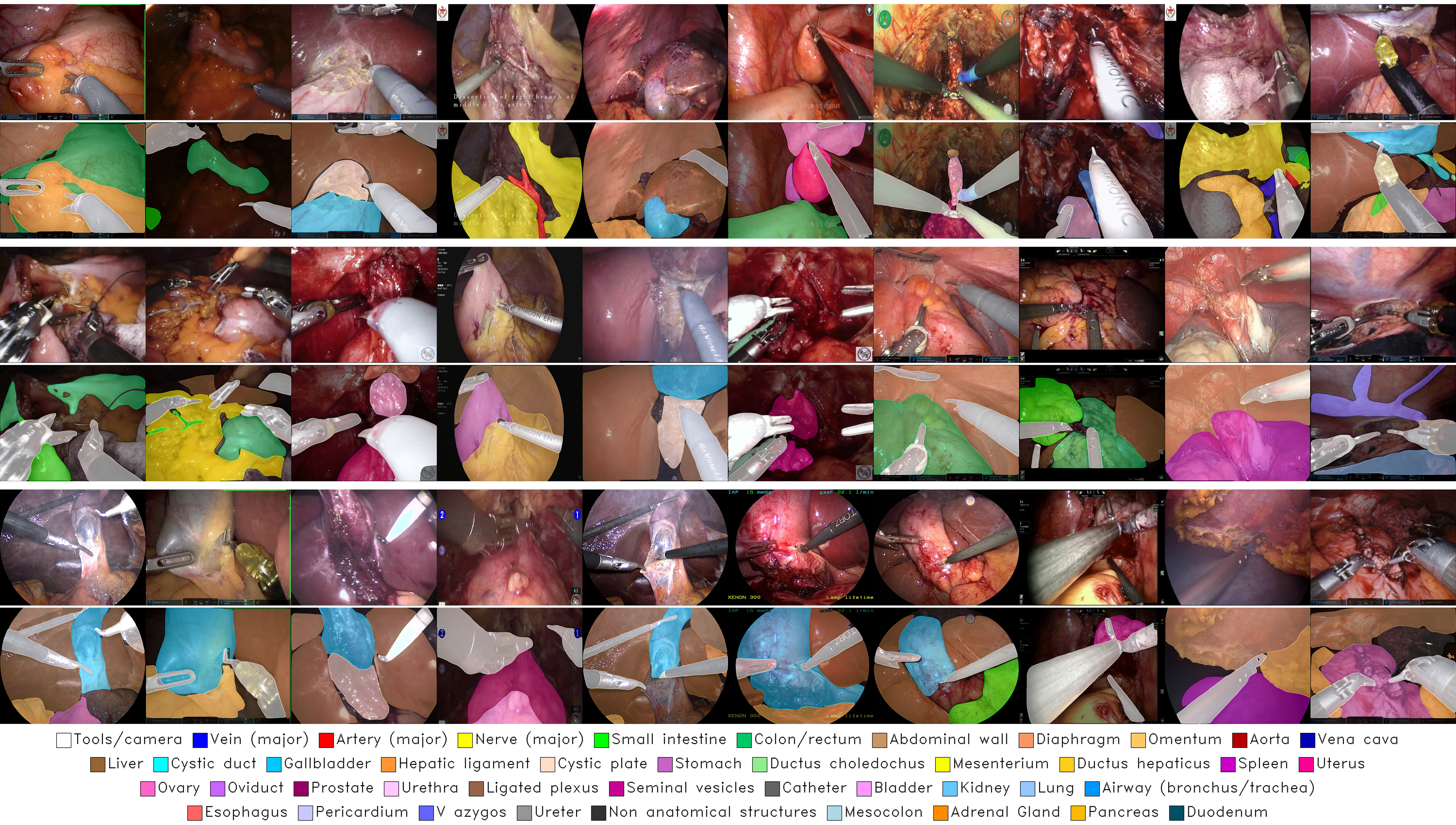}
    \caption{Examples of randomly selected frames and annotations, with at least one example from each included procedure and class.}
    \label{fig:examples}
\end{figure}

Figure~\ref{fig:examples} shows representative examples from the ATLAS-120k dataset, with at least one example per procedure. Table~\ref{tab:surgnetseg75k} compares ATLAS-120k with existing surgical datasets. ATLAS-120k distinguishes itself by including 14 distinct surgical procedures, a substantial expansion over prior datasets, which typically cover only a single procedure. Moreover, ATLAS-120k integrates both laparoscopic and robot-assisted minimally invasive surgical videos, thereby capturing a broader spectrum of procedural and technological variability. The Dresden Surgical Anatomy Dataset~(DSAD)~\cite{dsad} also contains a large number of frames annotated with multiple anatomical classes; however, most DSAD frames depict isolated anatomical segmentations rather than comprehensive semantic segmentation of the full surgical scene. Both the Endoscapes-Seg50~\cite{Mascagni2025} and CholecSeg8k~\cite{Cholecseg8k} datasets focus exclusively on minimally invasive surgery and contain fewer annotated frames than ATLAS-120k.

\begin{figure}[h]
    \centering
    \begin{subcaptionbox}{Distribution of clip durations.\label{fig:left}}[0.49\linewidth]
        {\includegraphics[width=\linewidth]{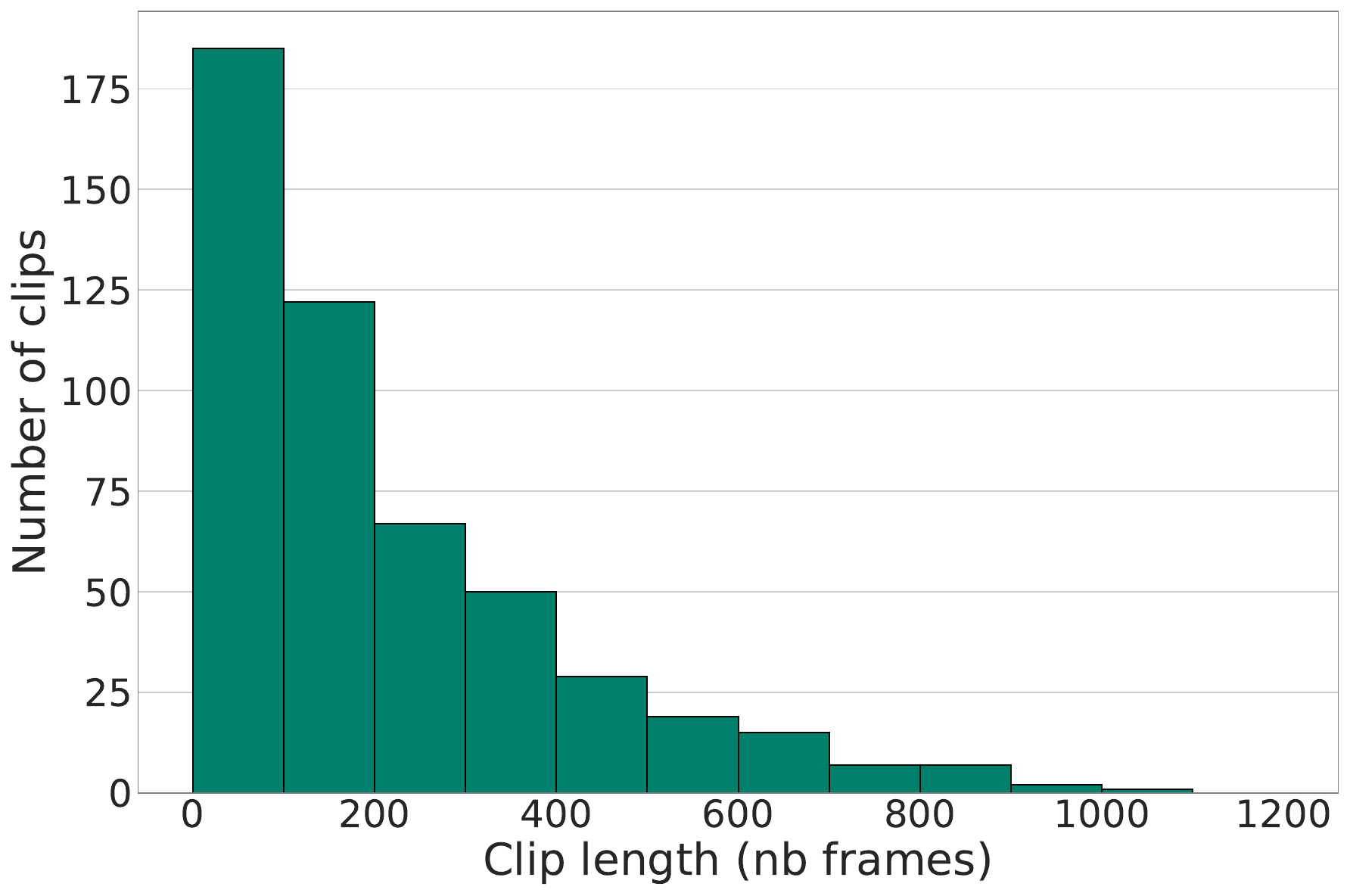}}
    \end{subcaptionbox}
    \hfill
    \begin{subcaptionbox}{Number of clips and frames per surgical procedure.\label{fig:right}}[0.49\linewidth]
        {\includegraphics[width=\linewidth]{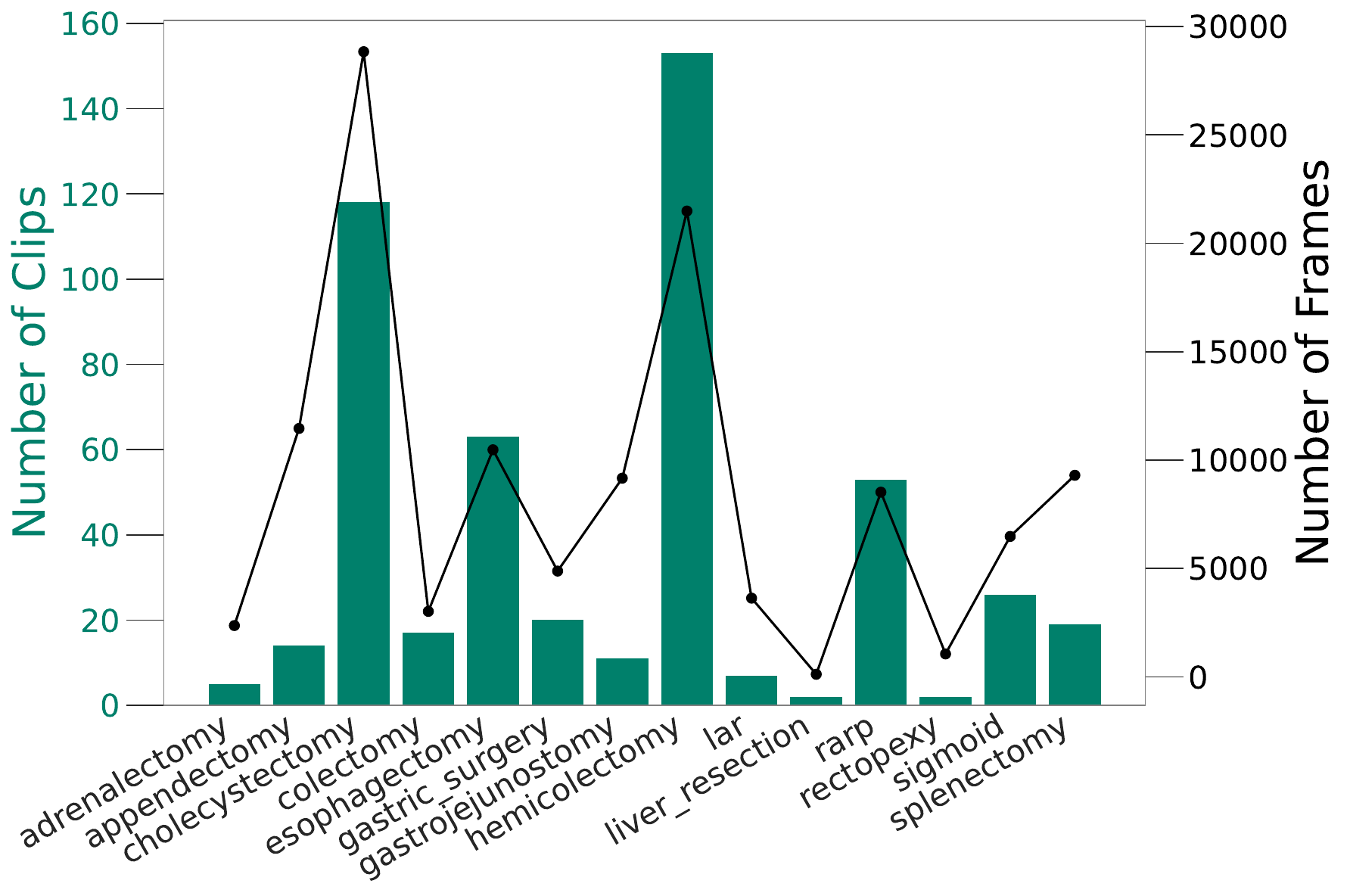}}
    \end{subcaptionbox}
    \caption{ATLAS-120k dataset characteristics.}
    \label{fig:data_distribution}
\end{figure}

A total of 100~videos were sourced from the GSViT dataset~\cite{schmidgall2024general}, from which 14 distinct procedures were selected for annotation. Rather than annotating full videos, clips of varying lengths were extracted. To standardize the dataset, all videos were downsampled to 15~fps, corresponding to the lowest original frame rate observed. Spatial resolutions varied from 480$\times$640 to 1080$\times$1920~pixels. Figure~\ref{fig:left} presents the distribution of clip durations, showing that most clips contain fewer than 400 frames ($<$30~seconds). Figure~\ref{fig:right} illustrates the number of clips and frames per procedure, with cholecystectomy contributing the largest share.

\begin{table*}
    \setlength{\tabcolsep}{4pt}
    \centering
    \caption{Comparison of surgical video segmentation datasets, summarizing key characteristics. MIS indicates minimally invasive surgery; RA indicates robot-assisted procedures.}
    \begin{tabular}{l | c c c c c c c }
        \hline
        \toprule
       \textbf{Dataset}           & \textbf{Included}  & \textbf{\#} & \textbf{\#}  & \textbf{\#} & \textbf{\#}  & \textbf{MIS}  & \textbf{RA} \\
       & \textbf{procedures} & \textbf{classes} & \textbf{videos} & \textbf{clips} & \textbf{frames} & & \\
        \midrule
        Endoscapes-Seg50~\cite{Mascagni2025}  & 1                    & 6          & 50          & -        & 493 & \checkmark &  \\
        CholecSeg8k~\cite{Cholecseg8k}       & 1                    & 12         & 17          & -        & 8,080 & \checkmark & \\   
        DSAD~\cite{dsad}              & 1                    & 11         & 32          & -       & 14,625 &  & \checkmark \\   
        ATLAS-120k    & 14                   & 42          & 100         & 502       & 121,018 & \checkmark & \checkmark \\
        \bottomrule
    \end{tabular}
    \vspace*{0.1cm}
    \label{tab:surgnetseg75k}
\end{table*}

Annotations were performed by three surgical research fellows under the supervision of three experienced surgeons (each with more than 10 years of experience). Although the primary objective was anatomical structure recognition, all visible surgical instruments were also annotated, enabling additional applications such as surgical tool segmentation.

The annotation workflow consisted of two phases. First, the initial frame of each clip was manually annotated with high precision. Second, a lightweight object-tracking model~\cite{cheng2023putting} was used to propagate these annotations across subsequent frames. Any propagation errors were manually corrected to ensure high annotation quality. To further improve tracking performance, the model was iteratively fine-tuned on the annotated data after approximately 10k, 25k, and 50k frames, using the original training parameters described in~\cite{cheng2023putting}.

Finally, the first frame of every clip was independently reviewed by at least one experienced surgeon to verify the correctness and completeness of the anatomical labels. The annotations of the dataset will be released as open source under a CC-BY-NC-SA-4.0 license to facilitate research in this direction.

\section{Experiments \& Results}
\textbf{Dataset.} In our experiments, the ATLAS-120k dataset was partitioned into patient-level training (70 videos), validation (10 videos), and test (20 videos) splits. To ensure consistency, we restricted model training to 30 classes by excluding categories that were not represented in all subsets and consolidating semantically similar classes. 

\textbf{Implementation details.} All our experiments were conducted on a single NVIDIA H100 GPU. The model was trained for 10 epochs with a batch size of 24. Each training sample consisted of a clip of length 5 frames. The architecture incorporated both procedural and temporal context representations, using one procedure context query and four phase queries. In addition, 100 segmentation queries were employed.
The training objective was formulated as a weighted sum of multiple loss components. Specifically, the mask binary cross-entropy loss and Dice loss were each assigned a coefficient of 5.0, the mask classification loss was weighted by 2.0, the procedure loss by 1.0, and the contrastive loss by 0.1. All hyperparameters were kept fixed across experiments unless otherwise stated. To improve model performance, the model was initialized with in-domain pretrained weights following the method described in~\cite{jong2026towards}. Data leakage was avoided by excluding ATLAS-120k from pretraining. 

\textbf{Evaluation.} All models are evaluated using detection and segmentation metrics. Detection performance is measured with COCO-style mean Average Precision (AP), including AP, AP75, and AP50, while segmentation quality is assessed using the Dice coefficient. Temporal stability is evaluated with the mean Video Consistency metric (mVC)~\cite{mvc}, computed over sliding windows of 12 and 24 consecutive frames. mVC quantifies the proportion of pixels whose predicted labels remain constant across a window, restricted to regions where ground-truth labels are temporally consistent and non-background. The final mVC score is obtained by averaging consistency across all valid windows and clips.

\textbf{State-of-the-art comparison.} Table~\ref{tab:results} shows the performance of ATLAS compared to state-of-the-art alternatives. Surgical foundation models such as SurgeNetXL~\cite{jaspers2025surgenet}, LEMON~\cite{che2025lemonlargeendoscopicmonocular}, and GSViT~\cite{schmidgall2024general} were excluded from this comparison due to overlap between their pretraining data and the ATLAS-120k dataset.  ATLAS consistently outperforms both natural-image and in-domain surgical foundation models across detection- and segmentation-oriented metrics. While DINO variants\cite{oquab2024dinov2learningrobustvisual, siméoni2025dinov3} show limited transfer to endoscopic videos, in-domain models such as EndoViT\cite{EndoVit} and SurgeNet improve performance but fall short of the strongest ATLAS variants. Our ViT-L model achieves 0.64 AP, 0.49 mDice, and 0.79 mVC24, demonstrating that combining foundation-model embeddings with temporal and procedural context queries yields substantial gains. Combined with an inference speed of 64 FPS (benchmarked on an NVIDIA H100), this enables real-time performance. Although the absolute mDice may appear moderate, it reflects the challenging nature of ATLAS-120k, which comprises 30 classes with highly imbalanced and low-prevalence anatomical structures. This class diversity and long-tail distribution make ATLAS-120k a realistic and demanding benchmark for future anatomy segmentation models.

\begin{table*}[t]
\setlength{\tabcolsep}{4pt}
\centering
\caption{Quantitative comparison of foundation models for segmentation. Models are grouped by pretraining domain: natural-image foundation models (top), in-domain pretrained medical models (middle), and our proposed ATLAS models (bottom).
For DINOv2 and DINOv3, we train a linear layer attached to the ViT. }
\resizebox{\columnwidth}{!}{%
\begin{tabular}{l l c | c c c c c c}
\toprule
Model & Backbone & \#PARAMS (M) $\downarrow$ & AP $\uparrow$ & AP75 $\uparrow$ & AP50 $\uparrow$ & mDice $\uparrow$ & mVC12 $\uparrow$ & mVC24 $\uparrow$ \\
\midrule
DINOv2~\cite{oquab2024dinov2learningrobustvisual} & ViT-B
& \cellcolor{green!20}87
& \cellcolor{red!10}0.32
& \cellcolor{red!8}0.29
& \cellcolor{red!10}0.61
& \cellcolor{red!25}0.09
& \cellcolor{green!10}0.69
& \cellcolor{green!5}0.57 \\

DINOv2~\cite{oquab2024dinov2learningrobustvisual} & ViT-L
& \cellcolor{red!10}304
& \cellcolor{red!15}0.29
& \cellcolor{red!18}0.25
& \cellcolor{red!10}0.61
& \cellcolor{red!22}0.10
& \cellcolor{red!5}0.61
& \cellcolor{red!5}0.49 \\

DINOv3~\cite{siméoni2025dinov3} & ViT-B
& \cellcolor{green!22}86
& \cellcolor{green!5}0.41
& \cellcolor{green!10}0.43
& \cellcolor{green!20}0.77
& \cellcolor{green!10}0.27
& \cellcolor{green!15}0.72
& \cellcolor{green!12}0.63 \\

DINOv3~\cite{siméoni2025dinov3} & ViT-L
& \cellcolor{red!12}303
& \cellcolor{green!8}0.43
& \cellcolor{green!12}0.44
& \cellcolor{green!15}0.74
& \cellcolor{green!12}0.28
& \cellcolor{green!12}0.70
& \cellcolor{green!10}0.60 \\

SAM2-UNet~\cite{SAM2UNet} & Hiera-L
& \cellcolor{red!5}217
& \cellcolor{green!10}0.45
& \cellcolor{green!15}0.46
& \cellcolor{green!18}0.76
& \cellcolor{green!5}0.24
& \cellcolor{red!2}0.63
& \cellcolor{red!2}0.52 \\

SAM3-UNet~\cite{SAM3UNet} & Hiera-L
& \cellcolor{red!25}449
& \cellcolor{green!18}0.48
& \cellcolor{green!22}0.51
& \cellcolor{green!18}0.76
& \cellcolor{green!20}0.35
& \cellcolor{green!18}0.73
& \cellcolor{green!15}0.64 \\

\midrule
EndoFM~\cite{endofm} & ViT-B
& \cellcolor{green!10}141
& \cellcolor{red!25}0.24
& \cellcolor{red!25}0.17
& \cellcolor{red!25}0.55
& \cellcolor{red!18}0.11
& \cellcolor{red!25}0.52
& \cellcolor{red!25}0.39 \\

GastroNet5M~\cite{GastroNet5m} & ViT-B
& \cellcolor{green!20}86
& \cellcolor{red!5}0.36
& \cellcolor{red!5}0.33
& \cellcolor{red!2}0.66
& \cellcolor{red!5}0.18
& \cellcolor{green!5}0.66
& \cellcolor{green!3}0.54 \\

EndoViT~\cite{EndoVit} & ViT-B
& \cellcolor{green!12}112
& \cellcolor{green!12}0.46
& \cellcolor{green!18}0.49
& \cellcolor{green!22}0.79
& \cellcolor{green!12}0.32
& \cellcolor{green!18}0.73
& \cellcolor{green!15}0.64 \\

SurgeNet-pvt~\cite{jaspers2025surgenet} & PVTv2
& \cellcolor{green!32}27
& \cellcolor{green!15}0.47
& \cellcolor{green!18}0.49
& \cellcolor{green!22}0.79
& \cellcolor{green!12}0.32
& \cellcolor{green!25}0.78
& \cellcolor{green!22}0.70 \\

SurgeNet-conv~\cite{jaspers2025surgenet} & ConvNeXtv2
& \cellcolor{green!30}30
& \cellcolor{green!22}0.50
& \cellcolor{green!25}0.52
& \cellcolor{green!20}0.78
& \cellcolor{green!10}0.31
& \cellcolor{green!5}0.66
& \cellcolor{green!3}0.54 \\

SurgeNet-ca~\cite{jaspers2025surgenet} & CaFormer-S18
& \cellcolor{green!33}26
& \cellcolor{green!25}0.51
& \cellcolor{green!30}0.55
& \cellcolor{green!25}0.80
& \cellcolor{green!12}0.32
& \cellcolor{red!5}0.61
& \cellcolor{red!5}0.49 \\

\midrule
ATLAS (ours) & ViT-S
& \cellcolor{green!35}24
& \cellcolor{green!28}0.52
& \cellcolor{green!28}0.54
& \cellcolor{green!25}0.80
& \cellcolor{green!10}0.31
& \cellcolor{green!20}0.75
& \cellcolor{green!18}0.66 \\

ATLAS (ours) & ViT-B
& \cellcolor{green!18}94
& \cellcolor{green!32}0.60
& \cellcolor{green!35}0.67
& \cellcolor{green!32}0.85
& \cellcolor{green!28}0.42
& \cellcolor{green!30}0.81
& \cellcolor{green!28}0.74 \\

ATLAS (ours) & ViT-L
& \cellcolor{red!8}318
& \cellcolor{green!35}0.64
& \cellcolor{green!35}0.70
& \cellcolor{green!35}0.89
& \cellcolor{green!35}0.49
& \cellcolor{green!35}0.85
& \cellcolor{green!35}0.79 \\

\bottomrule
\end{tabular}}
\label{tab:results}
\end{table*}

\textbf{Ablation study.} Table~\ref{tab:ablation_surgery} quantifies the contribution of each component in ATLAS. Starting from the EoMT baseline with default DINOv3 weights, in-domain pretraining using DINOv3 on the SurgeNet dataset~\cite{jaspers2025surgenet} provides a substantial boost in both detection and segmentation metrics. Adding temporal query propagation further improves performance, yielding higher AP and AP50. Incorporating context queries provides additional gains in detection, region-overlap, and temporal consistency metrics while introducing only minimal model complexity. Overall, these components improve AP by more than 20\% relative to the baseline while increasing the parameter count by less than 1\%. These results highlight that both temporal modeling and procedural context are critical for robust surgical video segmentation, a capability enabled by the scale and procedural diversity of ATLAS-120k. Additionally, Table~\ref{tab:ablation_surgery} shows the benefits of advanced pretraining methods for the ATLAS model.


\begin{table}[t]
\centering
\setlength{\tabcolsep}{4pt}
\caption{Ablation studies showing: (1) the incremental effect of tracking and context queries on surgical video segmentation, (2) the benefits of advanced pretraining.}
 \label{tab:ablation_surgery}
\resizebox{\columnwidth}{!}{%
\begin{tabular}{l c | c c c c c c}
\toprule
Method & \#PARAMS (M) $\downarrow$ & AP $\uparrow$ & AP75 $\uparrow$ & AP50 $\uparrow$ & mDice $\uparrow$ & mVC12 $\uparrow$ & mVC24 $\uparrow$ \\
\midrule
EoMT ViT-L (Baseline)
& 315
& \cellcolor{green!3}0.53
& \cellcolor{green!5}0.58
& \cellcolor{green!4}0.81
& \cellcolor{green!9}0.35
& \cellcolor{green!0}0.76
& \cellcolor{green!0}0.68 \\

$\rightarrow$ w/In-domain pretraining
& 315
& \cellcolor{green!15}0.57
& \cellcolor{green!15}0.62
& \cellcolor{green!19}0.85
& \cellcolor{green!24}0.43
& \cellcolor{green!16}0.80
& \cellcolor{green!19}0.74 \\

$\rightarrow$ w/Temporal query propagation
& 315
& \cellcolor{green!20}0.59
& \cellcolor{green!18}0.63
& \cellcolor{green!23}0.86
& \cellcolor{green!11}0.36
& \cellcolor{green!8}0.78
& \cellcolor{green!10}0.71 \\

$\rightarrow$ w/Context queries (ATLAS)
& 318
& \cellcolor{green!35}0.64
& \cellcolor{green!35}0.70
& \cellcolor{green!35}0.89
& \cellcolor{green!35}0.49
& \cellcolor{green!35}0.85
& \cellcolor{green!35}0.79 \\

\midrule

ATLAS ViT-B (DINOv1)
& 94
& \cellcolor{green!0}0.52
& \cellcolor{green!0}0.56
& \cellcolor{green!0}0.80
& \cellcolor{green!0}0.30
& \cellcolor{green!23}0.82
& \cellcolor{green!22}0.75 \\

$\rightarrow$ w/DINOv2
& 92
& \cellcolor{green!9}0.55
& \cellcolor{green!8}0.59
& \cellcolor{green!4}0.81
& \cellcolor{green!28}0.45
& \cellcolor{green!35}0.85
& \cellcolor{green!35}0.79 \\

$\rightarrow$ w/DINOv3
& 94
& \cellcolor{green!23}0.60
& \cellcolor{green!28}0.67
& \cellcolor{green!19}0.85
& \cellcolor{green!22}0.42
& \cellcolor{green!19}0.81
& \cellcolor{green!19}0.74 \\
\bottomrule
\end{tabular}}
\end{table}

\section{Conclusion}
We presented ATLAS-120k, a large-scale, diverse surgical video dataset, and ATLAS, a context-aware video segmentation model for anatomy recognition in minimally invasive surgery. By combining foundation-model embeddings with temporal and procedural context queries, ATLAS delivers accurate segmentation in real time. Experiments show that both temporal propagation and context modeling are essential for robust anatomy understanding. Together, the dataset and model provide a foundation for future research in anatomy-aware surgical perception and clinically relevant guidance systems.

\bibliographystyle{splncs04}
\bibliography{refs}

\end{document}